\documentclass[runningheads]{llncs}

\usepackage{eccv}

\usepackage{eccvabbrv}

\usepackage{graphicx}
\usepackage{booktabs}
\usepackage{enumitem}
\usepackage{amsmath,amssymb}
\usepackage{subcaption}
\usepackage{float}

\newcommand{\R}{\mathbb{R}}

\usepackage{graphicx}

\usepackage[accsupp]{axessibility}  

\usepackage{hyperref}

\usepackage{orcidlink}

\begin{document}

\title{Deep Networks Favor Simple Data} 


\author{Weyl Lu\inst{1} \and
Chenjie Hao\inst{1} \and
Yubei Chen\inst{1}}

\authorrunning{W.~Author et al.}

\institute{UC Davis, USA}

\maketitle

\begin{abstract}
Estimated density is often interpreted as indicating how typical a sample is under a model. Yet deep models trained on one dataset can assign \emph{higher} density to simpler out-of-distribution (OOD) data than to in-distribution test data. We refer to this behavior as the OOD anomaly. Prior work typically studies this phenomenon within a single architecture, detector, or benchmark, implicitly assuming certain canonical densities. We instead separate the trained network from the density estimator built from its representations or outputs. We introduce two estimators: Jacobian-based estimators and autoregressive self-estimators, making density analysis applicable to a wide range of models.

Applying this perspective to a range of models, including iGPT, PixelCNN++, Glow, score-based diffusion models, DINOv2, and I-JEPA, we find the same striking regularity that goes beyond the OOD anomaly: \textbf{lower-complexity samples receive higher estimated density, while higher-complexity samples receive lower estimated density}. This ranking appears within a test set and across OOD pairs such as CIFAR-10 and SVHN, and remains highly consistent across independently trained models. To quantify these rankings, we introduce Spearman rank correlation and find striking agreement both across models and with external complexity metrics. Even when trained only on the lowest-density (most complex) samples — or \textbf{even a single such sample} — the resulting models still rank simpler images as higher density.

These observations lead us beyond the original OOD anomaly to a more general conclusion: deep networks consistently favor simple data. Our goal is not to close this question, but to define and visualize it more clearly. We broaden its empirical scope and show that it appears across architectures, objectives, and density estimators.
\end{abstract}

\section{Introduction}

\paragraph{The Puzzle.}
Train a deep model on CIFAR-10, and ask it which images look more probable. One might expect the model to favor the data distribution it was trained on. Yet likelihood-based models have long displayed a disturbing habit: they can assign higher density to visually simpler out-of-distribution data, such as SVHN, than to the CIFAR-10 test images themselves \cite{nalisnick2019deep}. This behavior is often treated as a peculiarity of a particular detector, architecture, or benchmark. We believe that interpretation is too narrow. The real puzzle is not merely that likelihood sometimes fails as an OOD score. The deeper puzzle is that, across remarkably different deep networks, \emph{high estimated density keeps concentrating on simple data}.

\begin{figure}[H]
    \centering
    \makebox[\linewidth][c]{
        \includegraphics[width=\linewidth]{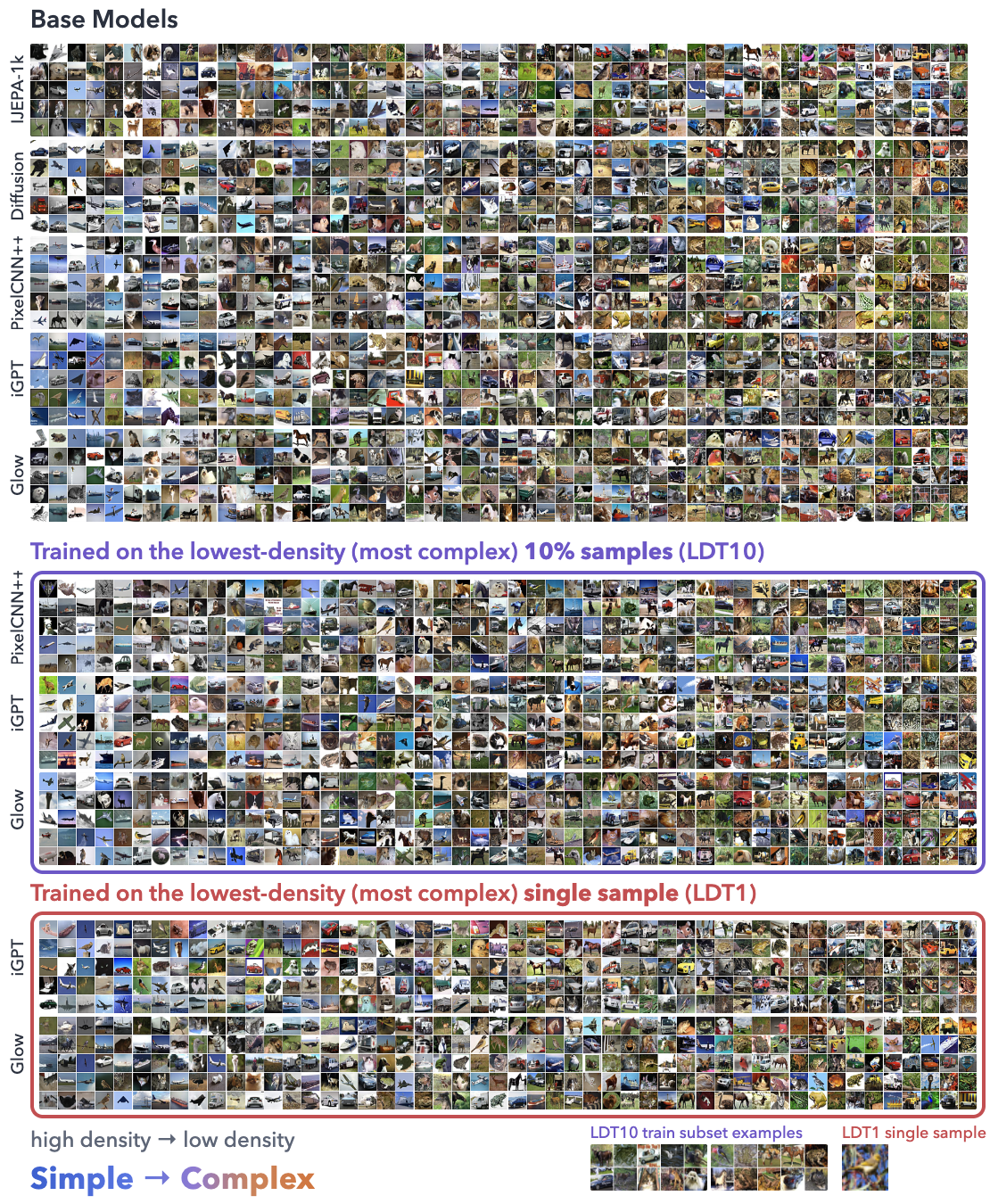}
    }
    \caption{\textbf{Density rankings on the CIFAR-10 test set.}
    For each model, CIFAR-10 test images are sorted by estimated density (high → low) and visualized through stratified samples along this ranking. 
    \textbf{Top:} base models trained on the full training set. 
    \textbf{Middle (LDT10):} models retrained on the \emph{lowest-density training subset} (the lowest 10\% of training samples). 
    \textbf{Bottom (LDT1):} models trained on a single lowest-density example. 
    Across all settings the ranking consistently progresses from \textbf{simple to complex images} (even under single-sample training). 
    All base models identify the same training image (CIFAR-10 id 29920) as the lowest-density sample; the corresponding LDT10 subset examples and the LDT1 example are shown on the right bottom.
    }
    \label{fig:hero}
\end{figure}

This paper starts from a simple empirical observation. If we rank CIFAR-10 test images from high to low estimated density, then across many independently trained deep models, the top-ranked images tend to be visually simple: smoother backgrounds, stronger low-frequency structure, larger homogeneous regions, and fewer intricate local details. The bottom-ranked images tend to be the opposite. More surprisingly, different models often induce almost the same ranking. Autoregressive image models, flow-based models, score-based diffusion models, and even representation learners frequently agree on which samples are ``easy'' and which are ``hard.'' The familiar CIFAR-10 versus SVHN effect is therefore not the whole story. It is only the most visible cross-dataset manifestation of a broader within-dataset ranking.

Existing work has illuminated important aspects of this problem without fully exhausting it. Nalisnick \etal\ showed that deep generative models can assign higher likelihood to OOD data and analyzed this effect through local second-order expansions around the data mean \cite{nalisnick2019deep}. Subsequent work explained parts of the phenomenon through architectural inductive bias in normalizing flows \cite{kirichenko2020why}, through the mismatch between density and typicality in high dimensions \cite{nalisnick2019typicality}, through input complexity and compression-based corrections \cite{serra2020input}, and through uncertainty-aware or ratio-based detectors such as model ensembles and likelihood ratios \cite{choi2018waic,ren2019likelihood,gangal2020likelihood}. These are valuable advances. But taken together, they still leave open a more basic question: \emph{why do deep networks, across architectures and training paradigms, keep assigning higher density to simpler samples in the first place?}

Our starting point is conceptual. A trained deep network should not be casually identified with one unique, canonical sample density. A network is a learned mapping; a density estimate is a statistical object constructed from that mapping. The distinction matters. Related work has studied several forms of simplicity bias in deep learning, including a bias toward simple input--output functions in function space \cite{valle2019simple}, pitfalls where SGD can over-rely on the simplest predictive features \cite{shah2020pitfalls}, and frequency/spectral biases where networks fit low-frequency components earlier or more readily \cite{rahaman2019spectral,xu2019frequency,belrose2024statistics}. These notions are important but not identical to the phenomenon we define here: our notion of simplicity is operationalized at the \emph{sample level} through the density rankings induced by network-based estimators. Once we separate the network from the density estimator built from it, many seemingly unrelated results fall into a common pattern. In this paper, we study \emph{network-induced density estimators}: ways of assigning a density score to an input by using either the model's explicit factorization or its learned feature geometry. This viewpoint is deliberately broader than the standard likelihood literature, and it lets us compare models that are usually discussed in separate communities.

We focus on two estimator families. The first is \emph{Jacobian-based estimation}. Here a sample is mapped to a latent or feature space equipped with a simple reference distribution, and its input-space density is estimated through a local Jacobian volume term; standard flow likelihood is the square, invertible special case, while more general feature maps yield a rectangular Jacobian estimator defined through singular-value volume correction \cite{rezende2020normalizing,caterini2021rectangular,balestriero2025gaussian}. This perspective also covers diffusion and score-based models through their continuous-flow interpretation, or equivalently through score integration \cite{song2021score}. The second is \emph{autoregressive self-estimation}, where the model directly factorizes the probability of a sequence into conditional probabilities over pixels or tokens, as in PixelCNN, iGPT, and GPT-style language models \cite{oord2016pixelcnn,radford2019gpt2,chen2020igpt}. One estimator is externally induced from the network's geometry; the other is internally provided by the model itself. Together, they give us a unified way to ask how different deep networks rank samples by density.

Viewed through these estimators, a striking empirical law emerges. Across all tested families, estimated density is systematically anti-correlated with sample complexity. The law appears \emph{within} a single in-distribution test set and does not require OOD data to be visible. It also appears \emph{across} datasets, where it recovers the classical OOD anomaly as a special case: simpler OOD datasets can outrank more complex in-distribution ones. This is not limited to a single architecture, modality, or training objective. We observe it in autoregressive image generators, flow-based models, score-based diffusion models, self-supervised representation learners such as DINOv2 and I-JEPA, and autoregressive language models.

The consistency of the induced rankings is one of the strongest pieces of evidence in our study. For CIFAR-10, independently trained models within the same family produce highly similar sample rankings, with strong Spearman and Kendall correlations across runs. This remains true not only for fully trained models but also under severe data restriction. If we first use a trained model to identify the lowest-density, most complex tail of the training set, and then retrain using only that 10\% subset, the new model still reconstructs a similar ranking over test samples. More strikingly, in our most extreme experiment, even training on a single low-density sample can still produce a nontrivial ranking that aligns with the broader simplicity ranking. These results suggest that the effect is not merely caused by the presence of many simple training examples. Rather, the ranking appears to reflect a deeper bias in how deep networks organize data once they are trained at all. We also find one notable caveat: DINOv2 is less consistent on CIFAR-10 than the other families, which we attribute to the strong mismatch between CIFAR-10 resolution and the much higher-resolution regime of DINOv2 pretraining.

What, then, is the right way to think about the classical OOD phenomenon? Our answer is that the OOD likelihood anomaly is not the main event. It is the visible tip of a larger regularity: \emph{deep networks prefer simple data}. In image space this preference aligns with low-frequency structure, reduced fine-scale variability, and lower external complexity measures; in text, it appears as high-likelihood but semantically impoverished or structurally repetitive strings. Once viewed in this way, many earlier ``fixes'' become easier to interpret. Methods based on likelihood ratios, compression-based background statistics, or model uncertainty often improve OOD detection precisely because they partially cancel or invert the underlying simplicity ranking, rather than because they fully recover a uniquely correct notion of semantic density.

Our goal in this paper is therefore modest in one sense and ambitious in another. We do not claim to provide a final theory of why this happens. But we do aim to define the phenomenon more clearly, broaden its empirical scope, and show that it is far more universal than the literature has typically acknowledged. Specifically, we make the following contributions:
\begin{itemize}
    \item We introduce a unified estimator viewpoint that separates a trained deep network from the density estimator induced by its outputs or learned features, allowing autoregressive, flow-based, diffusion/score-based, and representation-learning models to be studied in a common framework.
    \item We show that across all tested models, estimated density is consistently anti-correlated with sample complexity, both within in-distribution datasets and across classical in-distribution / out-of-distribution pairs.
    \item We demonstrate strong rank consistency across independently trained models using Spearman and Kendall correlations, and show that this consistency persists under severe data restriction, including retraining on only the lowest-density 10\% of the training set and, in an extreme setting, on a single low-density sample.
    \item We argue that the classical OOD likelihood anomaly should be understood not as an isolated failure mode of specific likelihood models, but as a special case of a broader simplicity preference of deep networks.
\end{itemize}

The rest of the paper develops this view. We first formalize the density estimators used throughout the paper. We then present the empirical phenomenology across models, datasets, and training regimes. Finally, we discuss why a preference for simple data may arise so broadly, and why existing OOD fixes may succeed without addressing the underlying phenomenon itself.

\section{Density Estimators, Rankings, and Complexity Measures}

This section introduces the three ingredients used throughout the analysis: density estimators, the rankings they induce, and external measures of image complexity.

Given a trained network $f_\theta$ and a density estimator $\mathcal E$, each image $x$ receives a score $s_{\theta,\mathcal E}(x)$. The analysis focuses on the \emph{ranking} induced by this score rather than on absolute density values. This density ranking is the primary observable in the paper: while the numerical scales of different estimators are often incomparable, the relative ranking of samples can be consistently compared across models, training settings, and external complexity measures.

\subsection{Density estimators}

We adopt a different viewpoint from much of the likelihood literature. 
Rather than assuming that each model comes with a single canonical density, 
the trained network and the density estimator derived from it are treated as two separate objects. 
A model provides representations or conditional predictions, while a density estimator is constructed from these quantities.

This separation makes it possible to analyze a wide range of architectures under a common framework. 
Two estimators are used throughout the paper: Jacobian-based estimators, which derive density-like scores from feature geometry, and autoregressive self-estimators, where the model directly factorizes the data likelihood.

\subsubsection{Jacobian-based estimators}

For invertible flow models, density is obtained exactly by change of variables. 
If $z=f_\theta(x)$ is an invertible square map and $p_0(z)$ is a tractable base density, then

\begin{equation}
\log \hat p(x)=\log p_0(f_\theta(x)) + \log |\det J_{f_\theta}(x)|.
\label{eq:flow}
\end{equation}

This is the standard flow likelihood \cite{kingma2018glow}.

The same geometric idea extends beyond square invertible maps. 
For a general representation $z=f_\theta(x)\in\R^m$, let $\sigma_1,\ldots,\sigma_r$ denote the nonzero singular values of the Jacobian. 
We use the local log-volume term

\begin{equation}
\log \operatorname{vol}(J_{f_\theta}(x))
=
\sum_{i=1}^{r}\log\sigma_i\!\bigl(J_{f_\theta}(x)\bigr),
\label{eq:logvol}
\end{equation}

which reduces to $\log|\det J|$ in the square case. 
Combined with a simple reference density in feature space (e.g., a Gaussian), this yields a Jacobian-based density estimator for representation models.

Balestriero et al.~\cite{balestriero2025gaussian} first applied this idea to a class of representation-learning models by assuming Gaussian feature embeddings and extracting density from the associated Jacobian geometry. 
Our use of the estimator is broader: we apply the same Jacobian-based construction to arbitrary networks, including models whose outputs or representations were not originally designed for density estimation (and even to models whose primary density estimator is autoregressive).

Moreover, the Gaussian assumption is not essential for the ranking behavior studied here. Empirically we find that a much weaker condition suffices: the variability of the feature-space log-density term is small compared to the Jacobian log-volume term. Under this regime the Jacobian contribution dominates the induced ranking. This observation will be discussed in more detail in Sec.~\ref{subsec:jacobian}.

For general encoders, the resulting quantity is treated as a principled density \emph{estimator} rather than necessarily a normalized density.

Score-based diffusion models admit density evaluation through the standard score-based likelihood route, equivalently through the probability-flow / score-integration formulation \cite{song2021score}. In the experiments, \emph{Diffusion} refers to the ImageNet-64 pretrained score-based diffusion model released with Dual Score Matching \cite{guth2025dual}.
When applied to CIFAR-10, high-resolution models such as DINOv2, I-JEPA, and Diffusion receive bicubically upsampled inputs before density estimation.

\subsubsection{Autoregressive self-estimators}

Autoregressive models provide an intrinsic density estimator through conditional factorization. 
For an image rasterized into a sequence $x=(x_1,\ldots,x_T)$,

\begin{equation}
\log \hat p_{\mathrm{AR}}(x)
=
\sum_{t=1}^{T}\log p_\theta(x_t\mid x_{<t}).
\label{eq:ar}
\end{equation}

This applies directly to autoregressive image models such as PixelCNN++ and iGPT \cite{salimans2017pixelcnnpp,chen2020igpt}. 
Unlike Jacobian-based estimators, the density score here is produced intrinsically by the model through next-pixel prediction.
\subsection{Ranking and rank correlation}

Given an evaluation set $\mathcal D=\{x_i\}_{i=1}^N$, each model--estimator pair induces a density ranking
$r_m(x_i)$ obtained by sorting images from highest to lowest estimated density. 
\textbf{This density ranking is the primary observable throughout the paper.}

To compare two rankings $r_a$ and $r_b$, we use Spearman rank correlation
\begin{equation}
\rho_s(a,b)=\operatorname{corr}\bigl(r_a(x_i),r_b(x_i)\bigr)_{i=1}^N .
\end{equation}

A value close to $1$ indicates that two methods produce nearly identical rankings, 
values near $0$ indicate weak agreement, and negative values indicate reversed rankings. 
Spearman correlation provides a common language for comparing density estimators, 
external complexity measures, and modified estimators throughout the paper.

Two correlation visualizations will be used repeatedly. 
In Fig.~\ref{fig:basecorr}, each lower-triangular entry shows the Spearman correlation between the full-dataset rankings produced by two models or proxies. 
In Fig.~\ref{fig:ldt}, each panel corresponds to one architecture family and compares rankings across training regimes: 
\emph{Base} uses the full CIFAR-10 training set, 
\emph{LDT10} retrains on the lowest-density $10\%$ subset of the training data, 
\emph{LDT1} retrains on the single lowest-density training image, 
and \emph{UT} denotes the randomly initialized untrained model.

\subsection{External complexity measures}

To relate density rankings to image complexity, we introduce two external proxies.

The first is \textbf{JPEG complexity}, defined as the negative compressed length of the JPEG representation so that larger values correspond to simpler images.

The second is a gradient-based proxy,
\begin{equation}
c_{\mathrm{grad}}(x)
=
-\log\!\Bigl(1+\operatorname{TV}(\mathrm{gray}(x))\Bigr),
\end{equation}
where $\operatorname{TV}(\mathrm{gray}(x))$ is the sum of the mean absolute horizontal and vertical differences of the grayscale image.

Both proxies are signed so that larger values correspond to simpler images. 
Positive Spearman correlation with these measures therefore indicates that a model ranks simpler images as higher density.
\section{Deep networks favor simple data}

\subsection{Base models reveal the same simple-to-complex axis}

The cleanest place to start is not OOD detection but within-dataset ranking. The \textbf{top row} of Fig.~\ref{fig:hero} sorts CIFAR-10 test images by estimated density for five base models. Across I-JEPA, Diffusion, iGPT, PixelCNN++, and Glow, the same visual progression appears again and again: higher-density images are smoother, cleaner, and more compressible, whereas lower-density images are busier, more textured, and compositionally irregular. This is already stronger than the classical OOD anomaly. Even inside one nominal test distribution, different model families rank samples along a similar simple-to-complex axis.

The \textbf{middle row} of Fig.~\ref{fig:hero} shows that this axis is not tied to training on the full dataset. After retraining on the lowest-density 10\% of CIFAR-10 training images (LDT10), iGPT, PixelCNN++, and Glow still rank evaluation images from simple to complex. The \textbf{bottom row} reveals an even more extreme case: after training on a single lowest-density image (LDT1), iGPT and Glow continue to produce a recognizable ranking over the full test set. In all three base-model families, the lowest-density CIFAR-10 training image selected by the base model is the \emph{same} training example, id 29920. This makes the single-sample result especially stark: the model is trained to overfit one complex image and still ends up preferring many unseen simple images.

\subsection{Interpreting the correlation maps}

\begin{figure}[t]
    \centering
    \includegraphics[width=\linewidth]{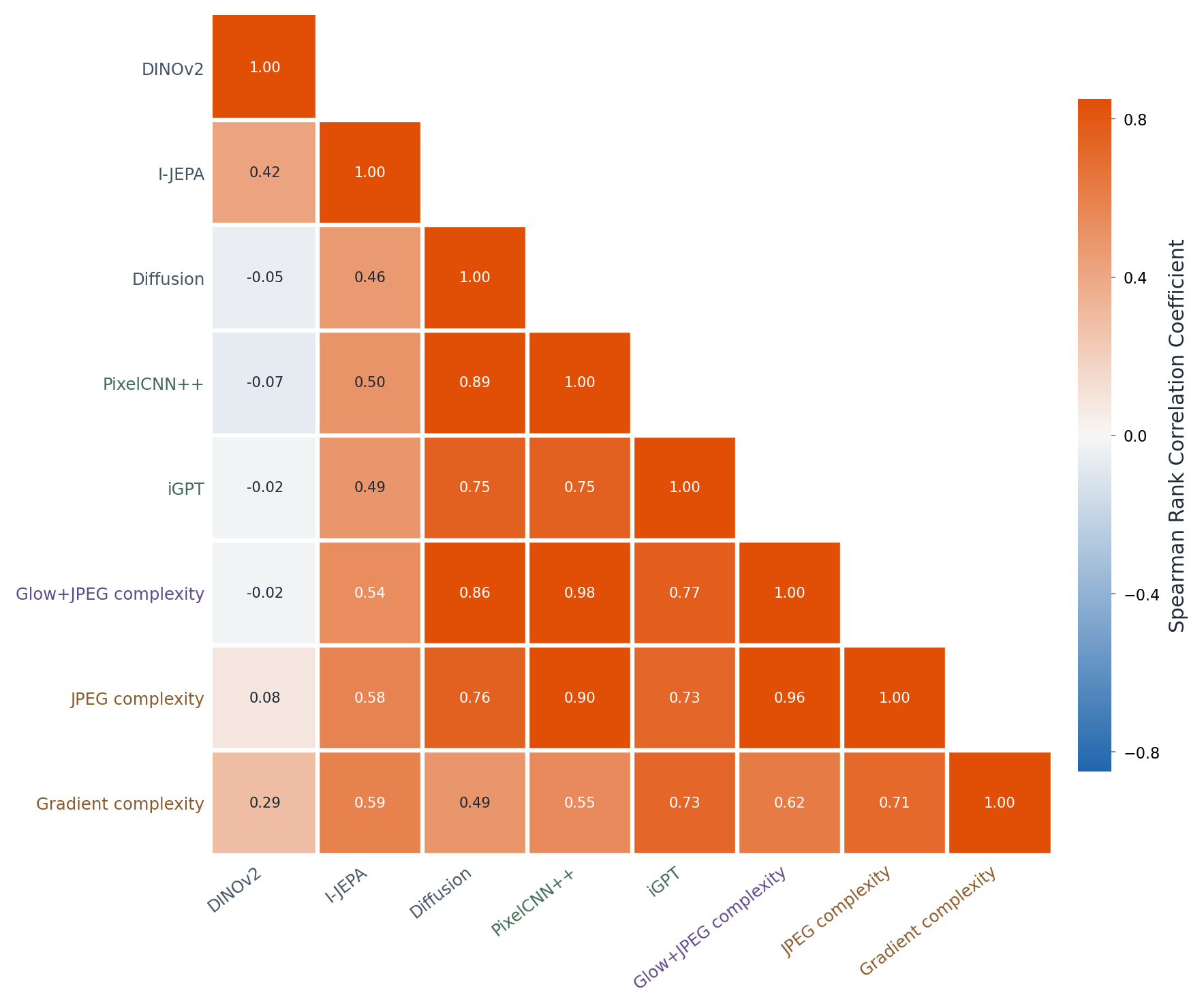}
    \caption{Spearman correlations between rankings induced by the base models and by external complexity proxies. The lower triangle is shown. Positive values mean two methods rank CIFAR-10 images in a similar rank from simple / high-density to complex / low-density. The row / column labeled ``Glow+JPEG complexity'' is the JPEG-based correction inspired by Serra et al. \cite{serra2020input}.}
    \label{fig:basecorr}
\end{figure}

Figure~\ref{fig:basecorr} summarizes the full-dataset ranking agreement. Read each entry as follows: pick a row and a column; the number in their intersection is the Spearman correlation between the two induced rankings. Thus, to ask whether two models sort CIFAR-10 in the same way, inspect their intersection. To ask whether a model follows an external notion of simplicity, inspect its intersection with JPEG or gradient complexity.

Several facts are immediate from Fig.~\ref{fig:basecorr}. First, excluding the DINOv2 row / column, the inter-model agreement is uniformly positive and often strong: I-JEPA correlates $0.46$ with Diffusion and $0.49$ with iGPT; Diffusion correlates $0.89$ with PixelCNN++ and $0.75$ with iGPT; PixelCNN++ and iGPT correlate $0.75$. These are substantial agreements between models with very different architectures, training objectives, and density estimators.

Second, the external complexity proxies line up with the model-induced rankings. Looking at the JPEG-complexity row / column in Fig.~\ref{fig:basecorr}, the correlations are $0.58$ with I-JEPA, $0.76$ with Diffusion, $0.90$ with PixelCNN++, and $0.73$ with iGPT. The gradient-complexity row / column shows the same trend, with correlations $0.59$, $0.49$, $0.55$, and $0.73$ respectively. These numbers should not be read as proving that JPEG compression or gradient total variation is the final answer. They are only proxies. What matters is that \emph{independently designed external complexity measures recover essentially the same sample ranking}.

Third, the JPEG-based correction inspired by Serra et al. \cite{serra2020input} does not eliminate the phenomenon. Inspect the ``Glow+JPEG complexity'' row / column in Fig.~\ref{fig:basecorr}: it still correlates $0.98$ with PixelCNN++, $0.86$ with Diffusion, $0.77$ with iGPT, and $0.96$ with JPEG complexity itself. In other words, the corrected estimator remains strongly aligned with the same underlying simple-to-complex direction. It can alter an OOD decision boundary without removing the bias.

The one conspicuous exception is DINOv2 on CIFAR-10. Its row / column is close to zero against Diffusion ($-0.05$), PixelCNN++ ($-0.07$), and iGPT ($-0.02$), and only moderately positive against I-JEPA ($0.42$). We return to this caveat in Sec.~\ref{subsec:dino}.

\subsection{Low-density retraining and single-sample training}

\begin{figure}[t]
    \centering
    \includegraphics[width=\linewidth]{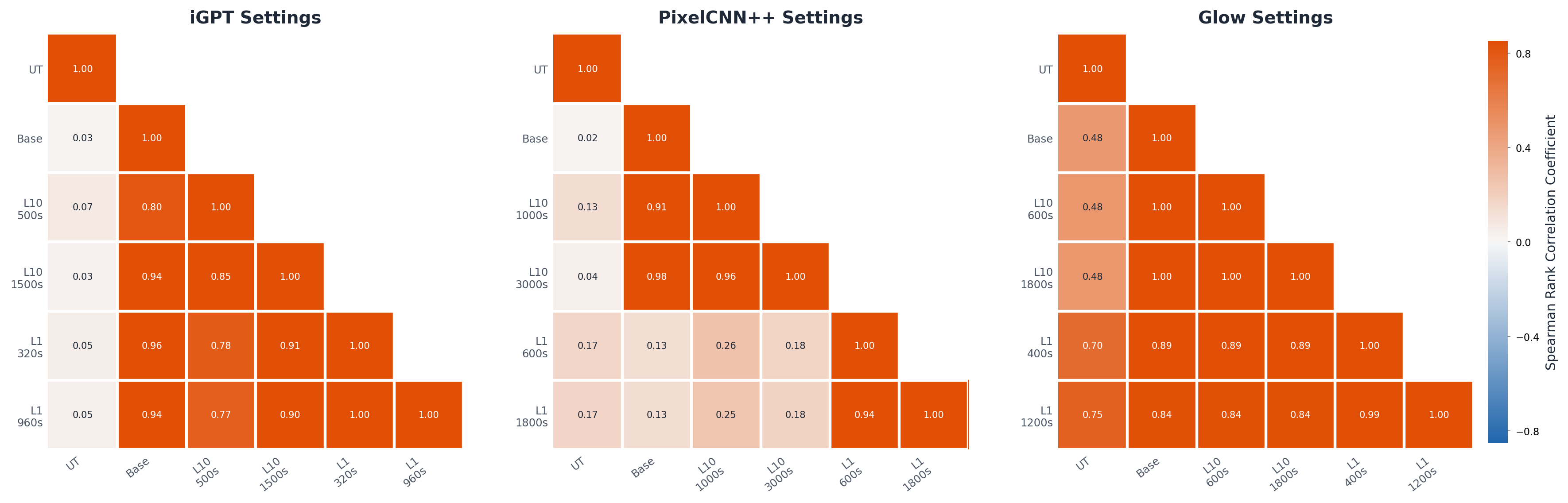}
    \caption{Spearman correlations between rankings induced by different training settings within each architecture family. Base uses the full CIFAR-10 training set, LDT10 uses only the lowest-density 10\% of the training set, LDT1 uses only the single lowest-density image, and UT is the untrained network. Each LDT setting is shown at two widely separated checkpoints to reduce the chance of reporting a transient undertrained state.}
    \label{fig:ldt}
\end{figure}

Figure~\ref{fig:ldt} is the key quantitative test of whether the ranking is merely inherited from the full dataset or actively regenerated by training. Each panel compares training settings within one architecture family. The first row / column of each panel compares the untrained model (UT) to trained checkpoints. The remaining entries compare Base, LDT10, and LDT1 rankings.

The left panel of Fig.~\ref{fig:ldt} shows that iGPT is remarkably stable. Base correlates $0.80$ and $0.94$ with the two LDT10 checkpoints, and $0.96$ and $0.94$ with the two LDT1 checkpoints. Thus, once iGPT has trained at all, even a single complex training image is enough to regenerate almost the same ranking axis.

The right panel shows the same phenomenon for Glow. Base correlates $1.00$ with both LDT10 checkpoints and still correlates $0.89$ and $0.84$ with the two LDT1 checkpoints. This is a striking result: the exact flow likelihood computed after single-image training continues to rank unseen CIFAR-10 images almost the same way as the full-dataset model.

\begin{figure}[t]
    \centering
    \includegraphics[width=\linewidth]{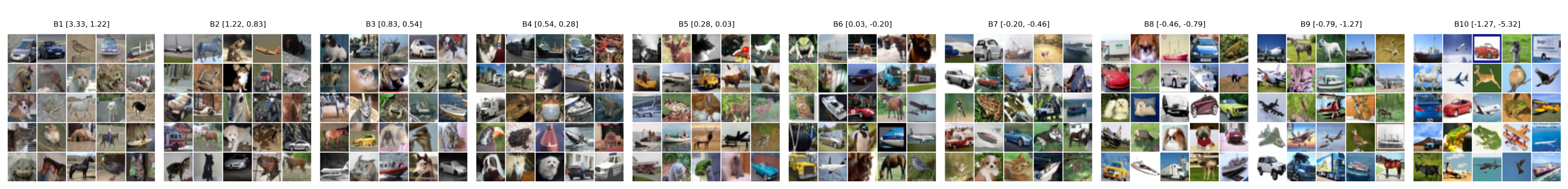}
    \caption{PixelCNN++ after training on a single lowest-density image. Unlike iGPT and Glow, the single-sample PixelCNN++ ranking no longer follows the global simplicity ranking from the base model. The failure is already visible quantitatively in the middle panel of Fig.~\ref{fig:ldt} and is shown qualitatively here.}
    \label{fig:pixelcnnldt1}
\end{figure}

PixelCNN++ is the interesting exception. In the middle panel, Base correlates $0.91$ and $0.98$ with the two LDT10 checkpoints, so the low-density-tail experiment still preserves the global ranking. But Base correlates only $0.13$ with each LDT1 checkpoint. Thus the single-sample regime breaks the ranking for PixelCNN++, even though the broader LDT10 result survives. We suspect that the convolutional autoregressive architecture is especially prone to overfitting local appearance statistics of the single image, particularly color and texture. This interpretation is consistent with the qualitative strip in Fig.~\ref{fig:pixelcnnldt1}, where the single-sample PixelCNN++ model ranks images more by superficial resemblance to the training image than by the broader simplicity axis.

The UT row / column in Fig.~\ref{fig:ldt} clarifies what is and is not innate. For iGPT and PixelCNN++, the untrained model is essentially uncorrelated with the trained ranking (Base correlations $0.03$ and $0.02$ respectively), so the preference is not present at initialization. Glow is different: the untrained Glow model already has a moderate correlation of $0.48$ with Base, and even higher correlations with the single-sample checkpoints ($0.70$ and $0.75$). Glow therefore seems to possess an architectural simplicity bias that training subsequently sharpens.

These results matter because they directly challenge a common tacit assumption: that a fitted density estimator primarily reflects the probability density of the training distribution. In the LDT10 and especially the LDT1 experiments, the model never sees the simple images it ends up preferring. No adversarial construction is needed. On natural data alone, the learned density ranking can be strongly decoupled from the empirical training distribution.

\subsection{Higher-resolution models and the DINOv2 caveat}
\label{subsec:dino}

The DINOv2 behavior on CIFAR-10 deserves special discussion. To score CIFAR-10 with DINOv2, I-JEPA, and Diffusion, we bicubically upsample the images to the corresponding input resolution. I-JEPA and Diffusion still exhibit the expected positive preference for simple images, but DINOv2 does not. This can already be seen in Fig.~\ref{fig:basecorr}, where the DINOv2 row / column is weakly correlated with most other models.

\begin{figure}[t]
    \centering
    \includegraphics[width=\linewidth]{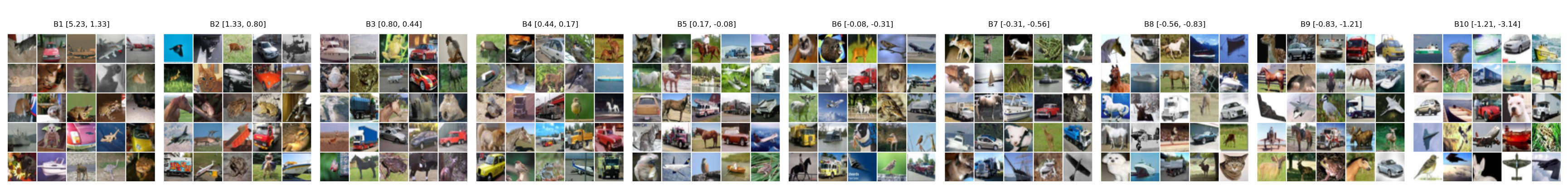}
    \caption{DINOv2 on bicubically upsampled CIFAR-10. Compared with the other models, the density bins display a much weaker monotone progression from simple to complex. This is consistent with the weak DINOv2 row / column in Fig.~\ref{fig:basecorr}.}
    \label{fig:dino}
\end{figure}

Figure~\ref{fig:dino} makes the CIFAR-specific DINOv2 behavior visible. Moving from the first density bin to the last does not produce the same monotone simple-to-complex progression seen in the other models. Our working interpretation is not that DINOv2 escapes the phenomenon in general, but that the combination of strong bicubic upsampling and DINOv2's own inductive bias disrupts the CIFAR-10 ranking.

\begin{figure}[t]
\centering

\begin{subfigure}[c]{0.69\linewidth}
    \centering
    \includegraphics[width=\linewidth]{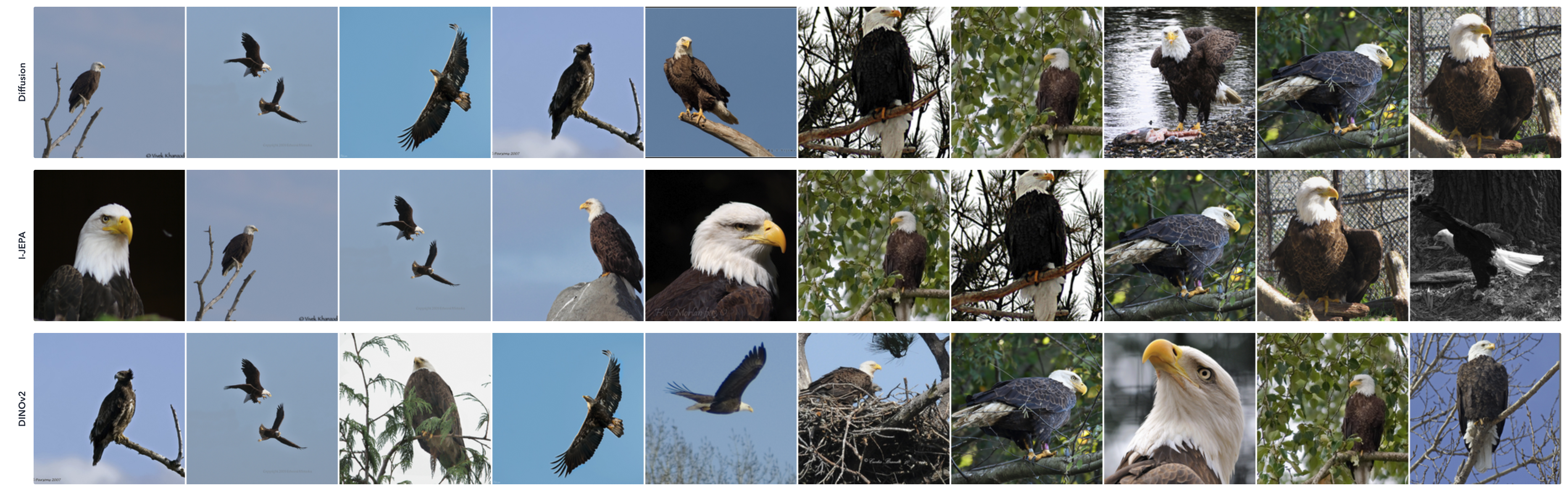}
    \caption{Highest- and lowest-density eagle images (ImageNet-1k/64) ranked by Diffusion, I-JEPA, and DINOv2. Across models, rankings exhibit a consistent simple-to-complex progression from left (high density) to right (low density).}
    \label{fig:imagenet_corr}
\end{subfigure}
\hfill
\begin{subfigure}[c]{0.30\linewidth}
    \centering
    \includegraphics[width=\linewidth]{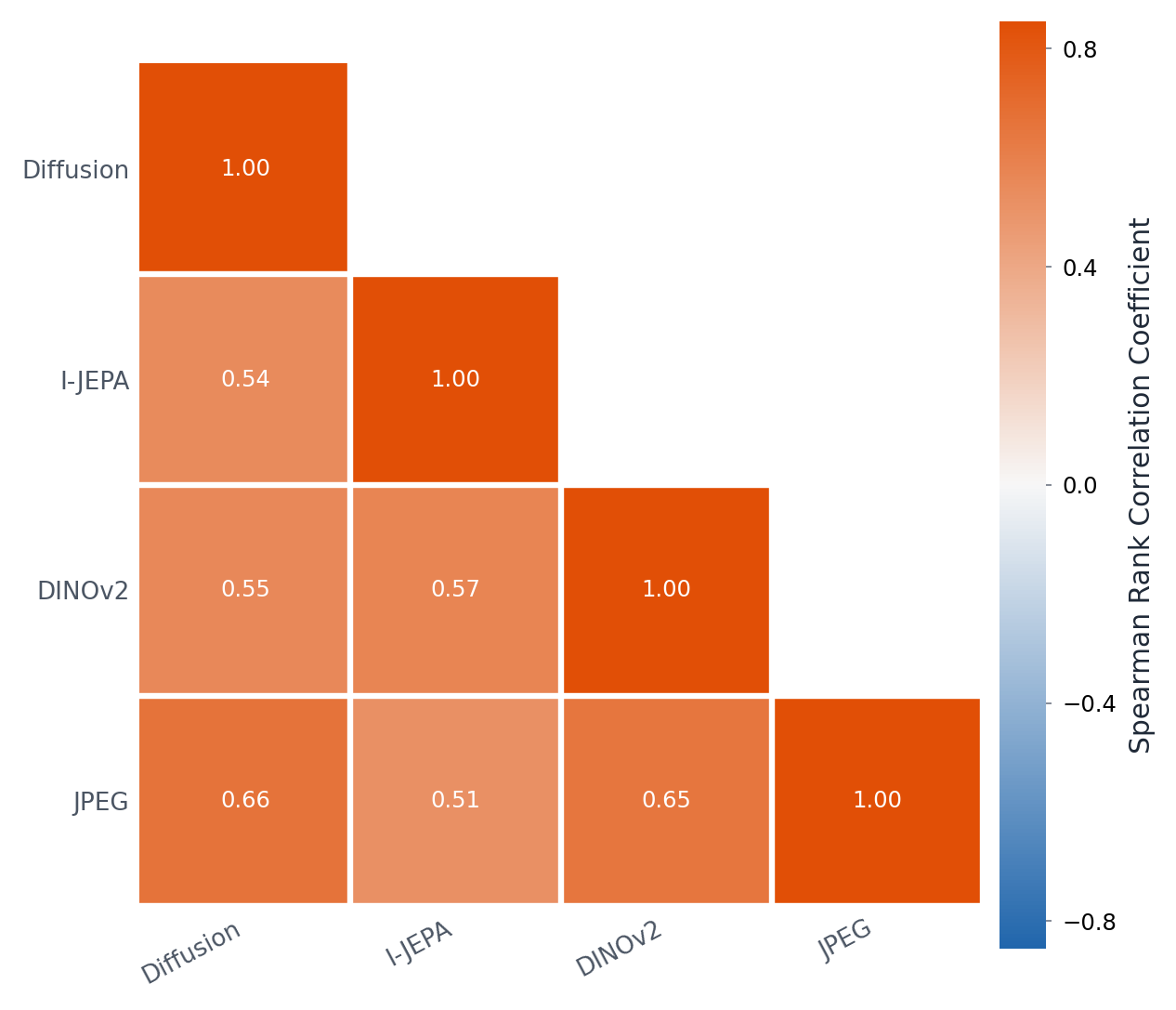}
    \caption{Spearman rank correlations between model rankings on the same eagle subset.}
    \label{fig:imagenet_vis}
\end{subfigure}

\caption{Density rankings on the ImageNet eagle subset. \textbf{Left:} qualitative rankings produced by Diffusion, I-JEPA, and DINOv2 show a consistent simple-to-complex ranking of images. \textbf{Right:} Spearman rank correlations confirm substantial agreement between the rankings produced by different models.}
\label{fig:eagle-highres}
\end{figure}

The higher-resolution analysis on ImageNet-1K matches the main trend. As shown in Fig.~\ref{fig:imagenet_corr}, Diffusion, I-JEPA, and DINOv2 exhibit clear positive inter-model rank agreement (Spearman $\rho_s\approx0.54$--$0.57$), and their rankings also align with JPEG-based complexity ($\rho_s\approx0.51$--$0.66$). Fig.~\ref{fig:imagenet_vis} further visualizes a single ImageNet-1K class: across all three models, high-density ranks concentrate on cleaner, simpler instances, while low-density ranks shift toward more cluttered and textured images. In contrast to the CIFAR-10 behavior in Fig.~\ref{fig:dino}, DINOv2 is not an outlier at ImageNet-1K resolution, supporting the interpretation that the CIFAR-10 discrepancy is driven by dataset--resolution interaction rather than a counterexample to the broader claim.

\subsection{Jacobian term dominates the ranking}
\label{subsec:jacobian}

\begin{figure}[t]
\centering

\begin{minipage}{0.49\linewidth}
    \centering
    \includegraphics[width=\linewidth]{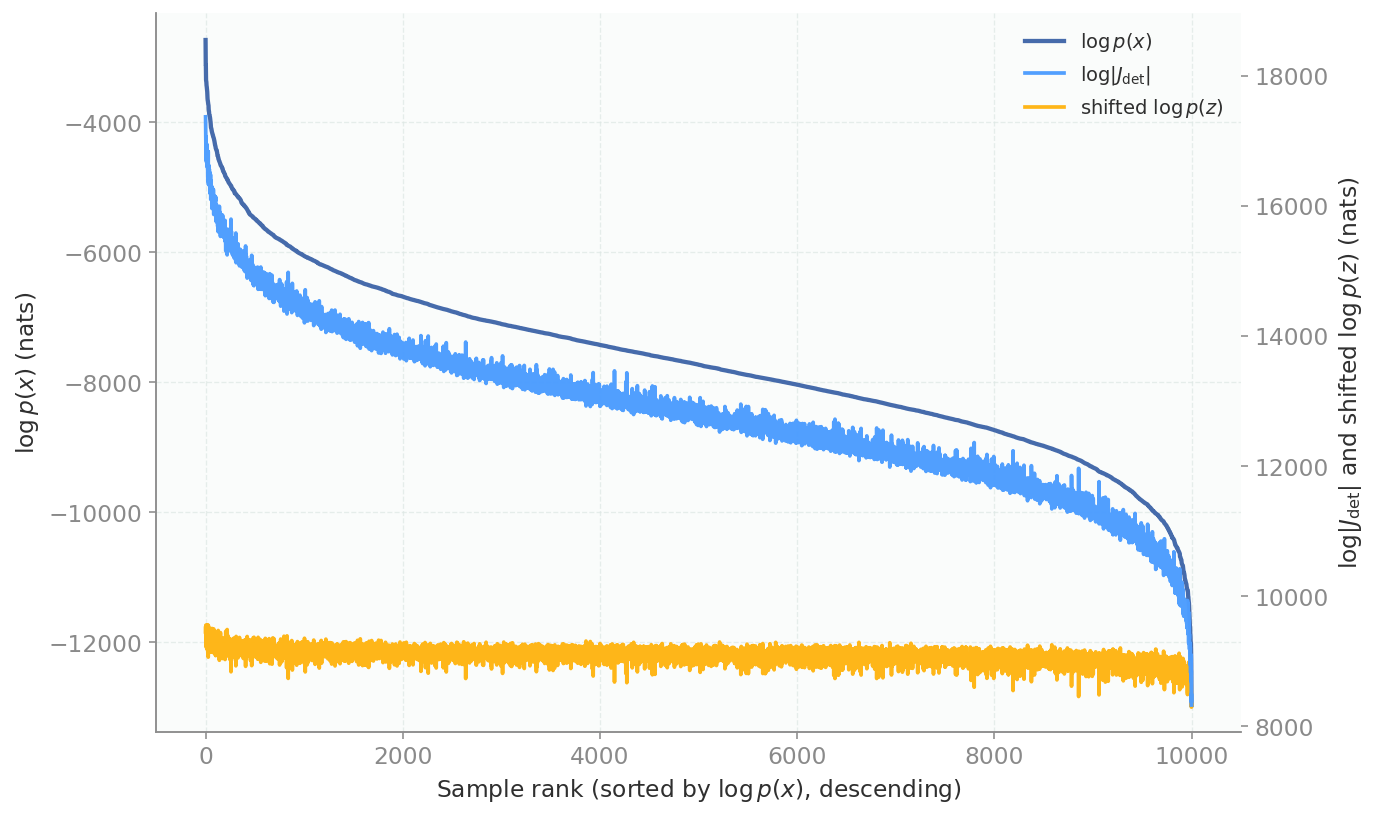}
    \caption{Glow likelihood decomposition on CIFAR-10, with samples sorted by $\log p(x)$ in descending order.}
    \label{fig:glowdom}
\end{minipage}
\hfill
\begin{minipage}{0.49\linewidth}
    \centering
    \includegraphics[width=\linewidth]{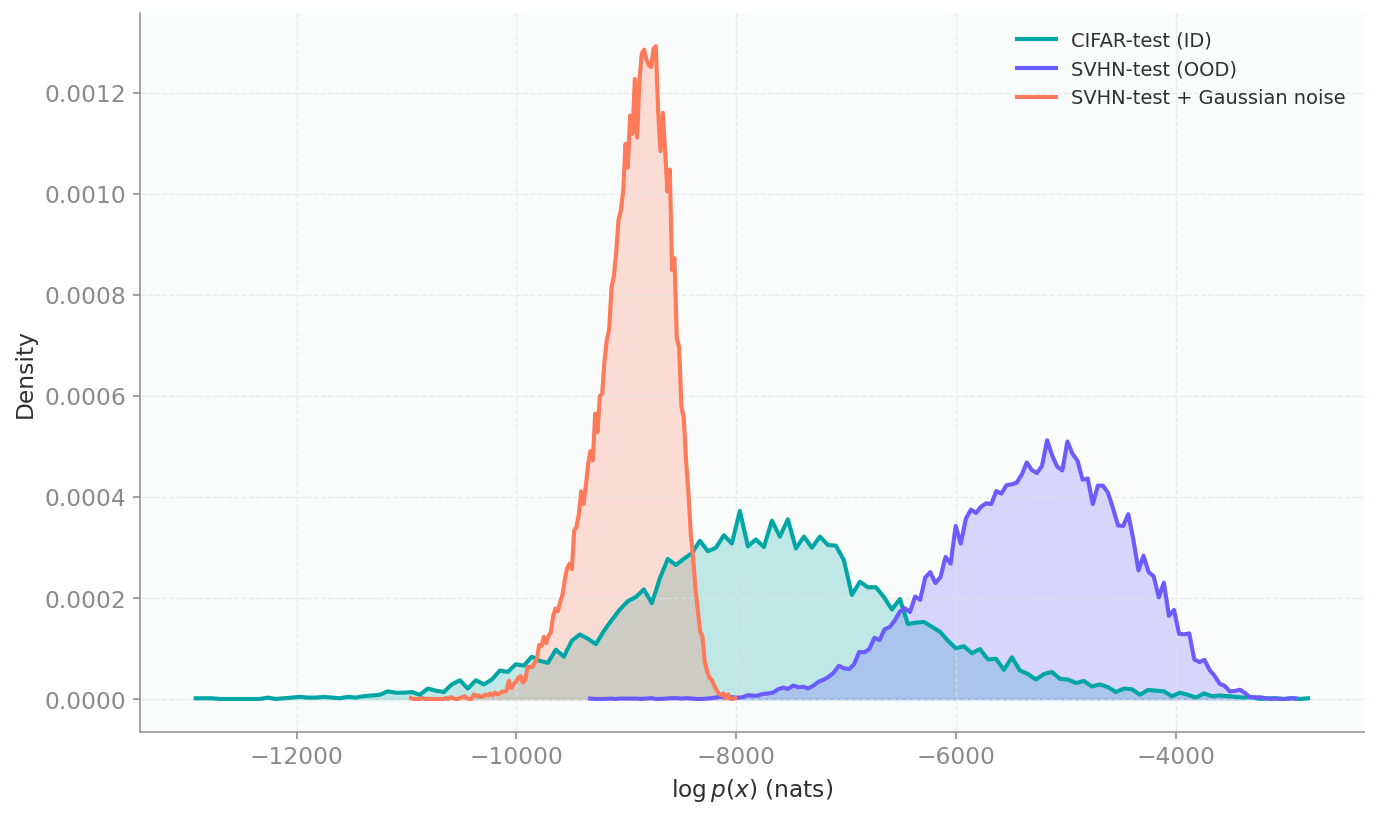}
    \caption{A tiny Gaussian perturbation destroys the classical OOD advantage of SVHN under Glow.}
    \label{fig:noise}
\end{minipage}

\end{figure}

For flow models, density decomposes into a base-density term and a Jacobian term as in Eq.~\eqref{eq:flow}. Figure~\ref{fig:glowdom} shows that, on Glow / CIFAR-10, the sample ranking is almost entirely controlled by the Jacobian term. Sorting images by $\log p(x)$ yields a Spearman correlation of $0.9975$ with $\log |J_{\mathrm{det}}|$, but only $0.5880$ with $\log p(z)$. The correlation between $\log p(z)$ and $\log |J_{\mathrm{det}}|$ is only $0.5366$.

Figure~\ref{fig:glowdom} also gives an intuition for why Jacobian-based estimators remain informative even when the reference density is only approximate. In high dimension, if the latent reference distribution is close to isotropic Gaussian, then $\log p(z)$ depends mainly on $\|z\|$ and varies relatively little because most samples lie on a thin shell. The Jacobian log-volume term is not subject to the same concentration effect and can dominate the ranking. We do not claim that this argument is universal or sufficient by itself, only that in our experiments the ranking of $\log p(x)$ is empirically much closer to the ranking of the Jacobian term than to the ranking of the latent reference term. This observation also helps motivate why JEPA-style Jacobian scores can work in practice even when the latent distribution is only approximately Gaussian \cite{balestriero2025gaussian}.

\subsection{Revisiting the OOD anomaly under small perturbations}

Figure~\ref{fig:noise} examines the robustness of the OOD likelihood ranking using a simple perturbation experiment. Glow trained on CIFAR-10 assigns higher likelihood to SVHN test images, reproducing the well-known phenomenon that models can prefer simpler out-of-distribution data. However, adding a very small Gaussian perturbation to SVHN ($\sigma^2=0.0064$) completely removes this advantage and shifts the noisy SVHN distribution below CIFAR-10.

To place this observation in context, recall that Nalisnick et al.~\cite{nalisnick2019deep} analyzed the original OOD anomaly using a second-order expansion of the log-density around a point $x_0$:

\[
\log p(x;\theta) \approx 
\log p(x_0;\theta)
+ \nabla_{x_0}\log p(x_0;\theta)^{\top}(x-x_0)
+ \tfrac12 (x-x_0)^{\top}
\nabla^2_{x_0}\log p(x_0;\theta)
(x-x_0).
\]

Comparing two data distributions $p^*$ and $q$ and taking expectations yields

\[
\mathbb{E}_q[\log p(x;\theta)]-
\mathbb{E}_{p^*}[\log p(x;\theta)]
\approx
\frac12
\mathrm{Tr}\!\left(
\nabla^2_{x_0}\log p(x_0;\theta)
(\Sigma_q-\Sigma_{p^*})
\right),
\]

which under diagonal approximations reduces to a comparison of channel-wise variances. 
Under this interpretation, datasets with smaller variance are expected to receive higher likelihood; since SVHN has lower channel-wise variance than CIFAR-10, the analysis predicts that SVHN should obtain higher expected log density.

Our perturbation experiment directly tests this explanation. 
The added noise barely changes the variance statistics emphasized above: SVHN pixel variance shifts only from $0.05023$ to $0.05027$, while CIFAR-10 remains substantially larger ($0.06262$), and the channel-wise ranking is unchanged. 
Yet the likelihood ranking reverses completely. 
Variance differences therefore cannot explain the anomaly: a local second-order expansion of the density around dataset means is far too coarse to capture the behavior of real data.
\section{Conclusion}

The evidence in this paper supports one central claim: \textbf{density estimators built from trained deep networks consistently favor simple data}. This claim is intentionally broader than any one architecture or benchmark. It covers intrinsic autoregressive likelihoods, exact flow likelihoods, score-based diffusion likelihoods, and Jacobian-based estimators on learned representations. It is visible within one dataset, across the classical CIFAR-10 / SVHN OOD pair, after retraining on only the lowest-density 10\% of the training set, and even after training on a single lowest-density sample.

Two implications deserve emphasis. First, the distinction between a trained network and the density estimator built from it is not merely philosophical. It is operationally necessary. Once the observable is the induced sample ranking rather than a single canonical density, models that are usually treated as incomparable fall into the same empirical pattern. Second, many existing ``fixes'' for OOD likelihood --- uncertainty correction, ratio-based correction, or external complexity adjustment --- should be interpreted with care. They can improve detection while leaving the underlying ranking largely intact.

We do not claim to have solved why this happens. Our most cautious reading is that the classical OOD likelihood anomaly is not the main phenomenon but only its most visible surface. The larger regularity is that, once trained, deep networks repeatedly allocate high estimated density to low-complexity images. Explaining why that ranking is regenerated across architectures, estimators, and even severely restricted training sets remains an open problem.

\clearpage  

%
%
\bibliographystyle{splncs04}
\bibliography{main}

@String(AAAI  = {AAAI})

@inproceedings{nalisnick2019deep,
  title = {Do Deep Generative Models Know What They Don't Know?},
  author = {Nalisnick, Eric and Matsukawa, Akihiro and Teh, Yee Whye and Gorur, Dilan and Lakshminarayanan, Balaji},
  booktitle = {International Conference on Learning Representations},
  year = {2019},
}

@misc{choi2018waic,
  title = {WAIC, but Why? Generative Ensembles for Robust Anomaly Detection},
  author = {Choi, Hyunsun and Jang, Eric and Alemi, Alexander A.},
  year = {2018},
  eprint = {1810.01392},
  archivePrefix = {arXiv},
  primaryClass = {stat.ML},
  doi = {10.48550/arXiv.1810.01392},
}

@misc{nalisnick2019typicality,
  title = {Detecting Out-of-Distribution Inputs to Deep Generative Models Using Typicality},
  author = {Nalisnick, Eric and Matsukawa, Akihiro and Teh, Yee Whye and Lakshminarayanan, Balaji},
  year = {2019},
  eprint = {1906.02994},
  archivePrefix = {arXiv},
  primaryClass = {stat.ML},
  doi = {10.48550/arXiv.1906.02994},
}

@inproceedings{ren2019likelihood,
  title = {Likelihood Ratios for Out-of-Distribution Detection},
  author = {Ren, Jie and Liu, Peter J. and Fertig, Emily and Snoek, Jasper and Poplin, Ryan and DePristo, Mark and Dillon, Joshua and Lakshminarayanan, Balaji},
  booktitle = {Advances in Neural Information Processing Systems},
  year = {2019},
}

@inproceedings{serra2020input,
  title = {Input Complexity and Out-of-Distribution Detection with Likelihood-Based Generative Models},
  author = {Serr\`a, Joan and {\'A}lvarez, David and G{\'o}mez, Vicen\c{c} and Slizovskaia, Olga and N{\'u}{\~n}ez, Jos{\'e} F. and Luque, Jordi},
  booktitle = {International Conference on Learning Representations},
  year = {2020},
}

@inproceedings{kirichenko2020why,
  title = {Why Normalizing Flows Fail to Detect Out-of-Distribution Data},
  author = {Kirichenko, Polina and Izmailov, Pavel and Wilson, Andrew Gordon},
  booktitle = {Advances in Neural Information Processing Systems},
  year = {2020},
}

@inproceedings{caterini2021rectangular,
  title = {Rectangular Flows for Manifold Learning},
  author = {Caterini, Anthony L. and Loaiza-Ganem, Gabriel and Pleiss, Geoff and Cunningham, John P.},
  booktitle = {Advances in Neural Information Processing Systems},
  year = {2021},
}

@inproceedings{song2021score,
  title = {Score-Based Generative Modeling through Stochastic Differential Equations},
  author = {Song, Yang and Sohl-Dickstein, Jascha and Kingma, Diederik P. and Kumar, Abhishek and Ermon, Stefano and Poole, Ben},
  booktitle = {International Conference on Learning Representations},
  year = {2021},
}

@misc{guth2025dual,
  title = {Learning Normalized Image Densities via Dual Score Matching},
  author = {Guth, Florentin and Kadkhodaie, Zahra and Simoncelli, Eero P.},
  year = {2025},
  eprint = {2506.05310},
  archivePrefix = {arXiv},
  primaryClass = {cs.LG},
  doi = {10.48550/arXiv.2506.05310},
}

@misc{balestriero2025gaussian,
  title = {Gaussian Embeddings: How JEPAs Secretly Learn Your Data Density},
  author = {Balestriero, Randall and Ballas, Nicolas and Rabbat, Mike and LeCun, Yann},
  year = {2025},
  eprint = {2510.05949},
  archivePrefix = {arXiv},
  primaryClass = {cs.LG},
  doi = {10.48550/arXiv.2510.05949},
}

@inproceedings{rahaman2019spectral,
  title = {On the Spectral Bias of Neural Networks},
  author = {Rahaman, Nasim and Baratin, Aristide and Arpit, Devansh and Draxler, Felix and Lin, Min and Hamprecht, Fred A. and Bengio, Yoshua and Courville, Aaron},
  booktitle = {International Conference on Machine Learning},
  year = {2019},
}

@inproceedings{shah2020pitfalls,
  title = {The Pitfalls of Simplicity Bias in Neural Networks},
  author = {Shah, Harshay and Tamuly, Kaustav and Raghunathan, Aditi and Jain, Prateek and Netrapalli, Praneeth},
  booktitle = {Advances in Neural Information Processing Systems},
  year = {2020},
}

@misc{belrose2024statistics,
  title = {Neural Networks Learn Statistics of Increasing Complexity},
  author = {Belrose, Nora and Pope, Quintin and Quirke, Lucia and Mallen, Alex and Dury, Brennan and Fern, Xiaoli},
  year = {2024},
  eprint = {2402.04362},
  archivePrefix = {arXiv},
  primaryClass = {cs.LG},
  doi = {10.48550/arXiv.2402.04362},
}

@inproceedings{kingma2018glow,
  title = {Glow: Generative Flow with Invertible 1x1 Convolutions},
  author = {Kingma, Diederik P. and Dhariwal, Prafulla},
  booktitle = {Advances in Neural Information Processing Systems},
  year = {2018},
}

@inproceedings{chen2020igpt,
  title = {Generative Pretraining from Pixels},
  author = {Chen, Mark and Radford, Alec and Child, Rewon and Wu, Jeffrey and Jun, Heewoo and Luan, David and Sutskever, Ilya},
  booktitle = {International Conference on Machine Learning},
  year = {2020},
}

@techreport{radford2019gpt2,
  title = {Language Models are Unsupervised Multitask Learners},
  author = {Radford, Alec and Wu, Jeffrey and Child, Rewon and Luan, David and Amodei, Dario and Sutskever, Ilya},
  institution = {OpenAI},
  year = {2019},
}

@inproceedings{valle2019simple,
  title = {Deep Learning Generalizes because the Parameter-Function Map is Biased Towards Simple Functions},
  author = {Valle-P{\'e}rez, Guillermo and Camargo, Chico Q. and Louis, Ard A.},
  booktitle = {International Conference on Learning Representations},
  year = {2019},
}

@inproceedings{gangal2020likelihood,
  title = {Likelihood Ratios and Generative Classifiers for Unsupervised Out-of-Domain Detection in Task-Oriented Dialog},
  author = {Gangal, Varun and Arora, Abhinav and Einolghozati, Arash and Gupta, Sonal},
  booktitle = {Proceedings of the AAAI Conference on Artificial Intelligence},
  year = {2020},
  pages = {7764--7771},
}

@inproceedings{oord2016pixelcnn,
  title = {Conditional Image Generation with {PixelCNN} Decoders},
  author = {van den Oord, Aaron and Kalchbrenner, Nal and Vinyals, Oriol and Espeholt, Lasse and Graves, Alex and Kavukcuoglu, Koray},
  booktitle = {Advances in Neural Information Processing Systems},
  year = {2016},
}

@inproceedings{salimans2017pixelcnnpp,
  title = {PixelCNN++: Improving the {PixelCNN} with Discretized Logistic Mixture Likelihood and Other Modifications},
  author = {Salimans, Tim and Karpathy, Andrej and Chen, Xi and Kingma, Diederik P.},
  booktitle = {International Conference on Learning Representations},
  year = {2017},
}

@misc{xu2019frequency,
  title = {Frequency Principle: {Fourier} Analysis Sheds Light on Deep Neural Networks},
  author = {Xu, Zhi-Qin John and Zhang, Yaoyu and Luo, Tao and Xiao, Yanyang and Ma, Zheng},
  year = {2019},
  eprint = {1901.06523},
  archivePrefix = {arXiv},
  primaryClass = {cs.LG},
  doi = {10.48550/arXiv.1901.06523},
}

@article{rezende2020normalizing,
  title = {Normalizing Flows for Probabilistic Modeling and Inference},
  author = {Papamakarios, George and Nalisnick, Eric and Rezende, Danilo Jimenez and Mohamed, Shakir and Lakshminarayanan, Balaji},
  journal = {Journal of Machine Learning Research},
  volume = {22},
  number = {57},
  pages = {1--64},
  year = {2021},
  url = {https://jmlr.org/papers/v22/19-1028.html},
  eprint = {1912.02762},
  archivePrefix = {arXiv},
  primaryClass = {stat.ML},
}
\end{document}